\documentclass{article}

\PassOptionsToPackage{numbers, compress}{natbib}

\usepackage[preprint]{neurips_2020}

\usepackage[utf8]{inputenc} %
\usepackage[T1]{fontenc}    %
\usepackage{hyperref}       %
\usepackage{url}            %
\usepackage{booktabs}       %
\usepackage{amsfonts}       %
\usepackage{nicefrac}       %
\usepackage{microtype}      %

\usepackage{subcaption}
\usepackage{graphicx} %
\graphicspath{{../images/}} %
\usepackage{amsmath}
\usepackage{colortbl}
\usepackage{multirow}
\usepackage{amssymb}
\usepackage{tablefootnote}
\usepackage{bm}
\usepackage{booktabs} %

\usepackage{wrapfig}
\usepackage{adjustbox}

\usepackage{xcolor}

\usepackage[normalem]{ulem}
\usepackage{cancel}

\allowdisplaybreaks

\newenvironment{rcases}
  {\left.\begin{aligned}}
  {\end{aligned}\right\rbrace}

\usepackage{xr}
\makeatletter
\newcommand*{\addFileDependency}[1]{
  \typeout{(#1)}
  \@addtofilelist{#1}
  \IfFileExists{#1}{}{\typeout{No file #1.}}
}
\makeatother

\title{Identifying Critical Neurons in ANN Architectures using Mixed Integer Programming}

\author{%
  Mostafa ElAraby \\
  Dept.\ of Comp.\ Sci.\ \& Oper.\ Res.,\\
  Mila, Universit\'e de Montr\'eal, \\
  Quebec, Canada \\
  \And
  Guy Wolf\thanks{equal contribution} \\
  Dept.\ of Math.\ \& Stat.,\\
  Mila, Universit\'e de Montr\'eal, \\
  Quebec, Canada \\
  \texttt{guy.wolf@umontreal.ca}\\
  \And
  Margarida Carvalho\textsuperscript{$*$} \\
  Dept.\ of Comp.\ Sci.\ \& Oper.\ Res.,\\
  CIRRELT, Universit\'e de Montr\'eal, \\
  Quebec, Canada \\
  \texttt{carvalho@iro.umontreal.ca}\\
}
\begin{document}
\maketitle

\newcommand{\myMatrix}[1]{\bm{\mathit{#1}}}

\begin{abstract}
We introduce a mixed integer program (MIP) for assigning importance scores to each neuron in deep neural network architectures which is guided by the impact of their simultaneous pruning on the main learning task of the network. 
By carefully devising the objective function of the MIP, we drive the solver to minimize the number of critical neurons (i.e., with high importance score) that need to be kept for maintaining the overall accuracy of the trained neural network.
Further, the proposed formulation generalizes the recently considered lottery ticket optimization by identifying multiple ``lucky'' sub-networks resulting in optimized architecture that not only performs well on a single dataset, but also generalizes across multiple ones upon retraining of network weights.
Finally, we present a scalable implementation of our method by decoupling the importance scores across layers using auxiliary networks.
We demonstrate the ability of our formulation to prune neural networks with marginal loss in accuracy and generalizability on popular datasets and architectures.
\end{abstract}

\section{Introduction}
\label{introduction}

Deep learning has proven its power to solve complex tasks and to achieve state-of-the-art results in various domains such as image classification, speech recognition,
machine translation, robotics and control \citep{bengio2017deep, lecun2015deep}.
Over-parameterized deep neural models with more parameters than the training samples can be used to achieve state-of-the art results on various tasks \citep{zhang2016understanding, neyshabur2018towards}.
However, the large number of parameters comes at the expense of computational cost in terms of memory footprint, training time and inference time on resource-limited  IOT devices \citep{lane2015early, li2018learning}.

In this context, pruning neurons from an over-parameterized neural model has been an active research area.
This remains a challenging open problem whose solution has the potential to increase computational efficiency and to uncover potential sub-networks that can be trained effectively.
Neural Network pruning techniques \citep{lecun1990optimal,hassibi1993optimal,han2015learning,srinivas2015data,dong2017learning,zeng2018mlprune,lee2018snip,wang2019eigendamage,salama2019pruning, serra2020lossless} have been introduced to sparsify models without loss of accuracy. Most existing work focus on identifying redundant parameters and non-critical neurons to achieve a lossless sparsification of the neural model. The typical sparsification procedure includes training a neural model, then computing parameters importance and pruning existing ones using certain criteria, and fine-tuning the neural model to regain its lost accuracy. Existing pruning and ranking procedures are computationally expensive, requiring iterations of fine-tuning on the sparsified model and no experiments were conducted to check the generalization of sparsified models across different datasets.

We remark that sparse neuron connectivity is often used by modern network architectures, and perhaps most notably in convolutional layers. Indeed, the limited size of the parameter space in such cases increases the effectiveness of network training and enables the learning of meaningful semantic features from the input images \citep{goodfellow2016deep}. Inspired by the benefits of sparsity in such architecture designs, we aim to leverage the neuron sparsity achieved by our framework to attain optimized neural architectures that can generalize well across different datasets.

\begin{figure}[ht]
    \centering
    \fbox{\includegraphics[width=0.6\textwidth]{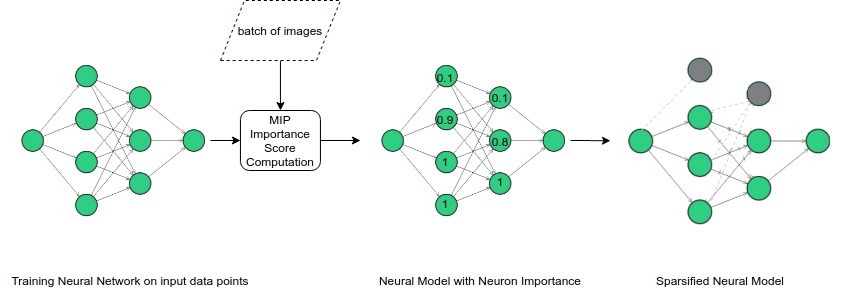}}
    \caption{The generic flow of our proposed framework used to remove neurons having an importance score less than certain threshold.}
    \label{fig:method_diagram}
\end{figure}
\paragraph{Contributions.}
In our proposed framework, illustrated in Figure~\ref{fig:method_diagram}, we formalize the notation of \emph{neuron importance} as a score between 0 and 1 for each neuron in a  neural network and the associated dataset.
The neuron importance score reflects how much activity decrease can be inflicted in it, while controlling the loss on the neural network model accuracy.
Concretely, we propose a mixed integer programming formulation (MIP) that allows the computation of each fully connected's neuron and convolutional feature map importance score and that takes into account the error propagation between the different layers. The motivation to use such approach comes from the existence of powerful techniques to solve MIPs efficiently in practice, and consequently, to allow the scalability of this procedure to large ReLU neural models. In addition, we extend the proposed formulation to support convolutional layers computed as matrices multiplication using toeplitz format~\citep{gray2000toeplitz} with an importance score associated with each feature map~\citep{molchanov2016pruning}.

Once neuron importance scores have been determined, a threshold is established to allow the identification of non-critical neurons and their consequent removal from the neural network. Since it is intractable to compute neuron scores using full datasets, in our experiments, we approximate their value by using subsets. In fact, we provide empirical evidence showing that our pruning process results in a marginal loss of accuracy (without fine-tuning) when the scores are approximated by a small balanced subset of data points, or even by parallelizing the scores' computation per class and averaging the obtained values.  Furthermore, we enhance our approach such that the importance scores computation is also efficient for very deep neural models, like VGG-16~\citep{simonyan2014very}. This stresses the scalability of it to datasets with many classes and large deep networks.
To add to our contribution, we show that the computed neuron importance scores from a specific dataset generalize to other datasets by retraining on the pruned sub-network, using the same initialization.

\paragraph{Organization of the paper. }
In Sec.~\ref{subsec:relatedwork}, we review relevant literature on neural networks sparsification and the use of mixed integer programming to model them. Sec.~\ref{subsec:background} provides background on the formulation of ReLU neural networks as MIPs. In Sec.~\ref{sec:mip_formulation}, we introduce the neuron importance score and its incorporation in the mathematical programing model, while Sec.~\ref{sec:mip_ojb} discusses the objective function that optimizes sparsification and balances accuracy.  Sec.~\ref{sec:experiments} provides computational experiments, and Sec.~\ref{sec:discussion}  summarizes our findings.

\subsection{Related Work}\label{subsec:relatedwork}
\paragraph{Classical weight pruning methods.} \citet{lecun1990optimal} proposed the optimal brain damage that theoretically prunes weights having a small saliency by computing its second derivatives with respect to the objective. The objective being optimized was the model's complexity and training error. \citet{hassibi1993second} introduced the optimal brain surgeon that aims at removing non-critical weights determined by the Hessian computation. Another approach is presented by \citet{chauvin1989back, weigend1991generalization}, where a penalizing term  is added to the loss function during the model training (e.g.\ L0 or L1 norm) as a regularizer. The model is sparsified during backpropagation of the loss function.  Since these classical methods depend \emph{i)} on the scale of the weights, \emph{ii)} are incorporated during the learning process, and, some of them, \emph{iii)} rely on computing the Hessian with respect to some objective, they turn out to be slow, requiring iterations of pruning and fine-tuning to avoid loss of accuracy. On the other hand, our approach identifies a set of non-critical neurons that when pruned simultaneously results in a marginal loss of accuracy without the need of fine-tuning or re-training.

\paragraph{Weight pruning methods.} \citet{molchanov2016pruning} devised a greedy criteria-based pruning with fine-tuning by backpropagation. The criteria devised is given by the absolute difference between dense and sparse neural model loss (ranker). This cost function ensures that the model will not significantly decrease its predictive capacity. The drawback of this approach is in requiring a retraining after each pruning step. \citet{shrikumar2017learning} developed a framework that computes the neurons' importance at each layer through a single backward pass. This technique compares the activation values among the  neurons and assigns a contribution score to each of them based on each input data point. Other related techniques, using different objectives and interpretations of neurons importance, have been presented \citep{berglund2015measuring,barros2018crosstalk,liu2018understanding,yu2018nisp,hooker2019benchmark,srinivas2015data,jordao2018pruning,he2018soft}. They all demand intensive computation, while our approach aims at efficiently computing neurons' importance. \citet{lee2018snip} investigates the pruning of connections, instead of entire neurons. The connections' sensitivity is studied through the model's initialization and a batch of input data. The sensitivity of the connections are computed using the magnitude of the derivatives of the mini-batch with respect to the loss function. Connections having a sensitivity score lower than a certain threshold are removed. This proposed technique is the current state-of-the-art in deep networks' compression.
\paragraph{Lottery ticket.} \citet{frankle2018lottery} introduced the lottery ticket theory that shows the existence of a lucky pruned sub-network,  a \emph{winning ticket}. This winning ticket can be trained effectively with fewer parameters, while achieving a marginal loss in accuracy. \citet{morcos2019one} proposed a technique for sparsifying \(n\) over-parameterized trained neural model based on the lottery hypothesis. Their technique involves pruning the model and disabling some of its sub-networks. The pruned model can be fine-tuned on a different dataset achieving good results. To this end, the dataset used for on the pruning phase needs to be large.  The lucky sub-network is found by iteratively pruning the lowest magnitude weights and retraining. Another phenomenon discovered by \citep{ramanujan2020s, wang2020pruning}, was the existence of smaller high-accuracy models that resides within larger random networks. This phenomenon is called strong lottery ticket hypothesis and was proved by~\citep{malach2020proving} on ReLU fully connected layers. Furthermore, \citet{wang2020picking} proposed a technique of selecting the winning ticket at initialization before training the ANN by computing an importance score, based on the gradient flow in each unit.

\paragraph{Mixed-integer programming} \citet{fischetti2018deep} and \citet{anderson2019strong} represent a ReLU ANN using a MIP\@. \citet{fischetti2018deep} presented a big-M formulation to represent trained ReLU neural networks. Later, \citet{anderson2019strong} introduced the strongest possible tightening to the big-M formulation by adding strengthening separation constraints when needed\footnote{ The cut callbacks in Gurobi were used to inject separated inequalities into the cut loop.}, which reduced the solving time by orders of magnitude. All the proposed formulations, are designed to represent trained ReLU ANNs with fixed parameters. In our framework, we used the formulation from \citep{fischetti2018deep} because its performance was good due to  our tight local variable bounds, and its polynomial number of constraints (while, \citet{anderson2019strong}'s model has an exponential number of constraints). Representing ANN as a MIP can be used to evaluate robustness, compress networks and create adversarial examples for the trained neural network. \citet{tjeng2017evaluating} used a big-M formulation to evaluate the robustness of neural models against adversarial attacks. In their proposed technique, they assessed the ANN's sensitivity to perturbations in input images. The MIP solver tries to find a perturbed image (adversarial attack) that would get misclassified by the ANN\@. \citet{serra2020lossless} also used a MIP to maximize the compression of an existing neural network without any loss of accuracy. Different ways of compressing (removing neurons, folding layers, etc) are presented. However, the reported computational experiments lead only to the removal of inactive neurons. Our method has the capability to identify such neurons, as well as to identify other units that would not significantly compromise accuracy.

\citet{huang2019achieving} used also mathematical programming models to check neural models' robustness in the domain of natural language processing. In their proposed technique, the bounds computed for each layer would get shifted by an epsilon value for each input data point for the MIP\@. This epsilon is the amount of expected perturbation in the input adversarial data.

\subsection{Background and Preliminaries}\label{subsec:background}
\emph{Integer programs} are combinatorial optimization problems restricted to discrete variables, linear constraints and linear objective function. These problems are NP-hard, even when variables are restricted to be binary~\citep{Garey:1979}. The difficulty comes from ensuring integer solutions, and thus, the impossibility of using gradient methods. When continuous variables are included, they are designated by \emph{mixed integer programs}. Advances in combinatorial optimization such as branching techniques, bounds tightening, valid inequalities, decomposition and heuristics, to name few, have resulted in powerful solvers that can in practice solve MIPs of large size in seconds. See~\citep{Wolsey1988} for an introduction to integer programming.

Consider layer \(l\) of a trained ReLU neural network with \(\myMatrix{W^l}\) as the weight matrix, \(w_i^l\) row \(i\) of  \(\myMatrix{W^l}\), and  \(b^l\) the bias vector. For each input data point \(x\),  let \(h^l\)  be a decision vector denoting the output value of layer \(l\), i.e. \(h^l = ReLU(\myMatrix{W^l} h^{l-1}+b^l)\) for \(l>0\) and \(h^{0}=x\), and \(z_i^l\) be a binary variable taking value 1 if the unit \(i\) is active, i.e. \(w_i^l h^{l-1}+b^l_i \geq 0\), and \(0\) otherwise. Finally, let \(L^{l}_{i}\) and \(U^{l}_{i}\) be constants indicating a valid lower and upper bound for the input of each neuron \(i\) in layer \(l\). We discuss the computation of these bounds in Sec.~\ref{sec:bound_prop}. For now, we assume that \(L^{l}_{i}\) and \(U^{l}_{i}\) are  sufficiently  small and  large numbers, respectively, i.e., the so-called Big-M values. Next, we provide the standard constraint representation of ReLU neural networks. For sake of simplicity, we describe the formulation for one layer \(l\) of the model at neuron \(i\) and one input data point \(x\):
\begin{subequations}
    \begin{alignat}{4}
        h^{0}_i   = x_i                                                     &  & \quad \textrm{if } l=0,  \textrm{ otherwise }\label{eq:equality_input} \\
        h_{i}^l   \geq 0,                                                   &  & \label{eq:always_positive}                                             \\
        h^{l}_{i} + (1- z^{l}_i) L^{l}_i   \leq w_i^{l} h^{l-1} + b^{l}_i , &  & \label{eq:z_i_upper}                                                   \\
        h^{l}_{i}   \leq z^{l}_i U^{l}_i,                                   &  & \label{eq:upper_bound_relu}                                            \\
        h^{l}_{i}  \geq w_i^{l} h^{l-1} + b^{l}_i,                          &  & \label{eq:z_i_lower}                                                   \\
        z^{l}_{i} \in \{0, 1\}.                                             &  & \label{eq:v_bounds_1}
    \end{alignat}
    \label{eq:relu_constraints}
\end{subequations}

In~\eqref{eq:equality_input}, the initial decision vector \(h^{0}\) is forced to be equal to the input \(x\) of the first layer. When \(z^{l}_i\) is 0,  constraints~\eqref{eq:always_positive} and~\eqref{eq:upper_bound_relu} force \(h^l_i\) to be zero, reflecting a non-active neuron. If an entry of \(z^{l}_i\) is 1, then constraints~\eqref{eq:z_i_upper} and~\eqref{eq:z_i_lower} enforce  \(h^l_i\) to be equal to \(w_i^{l} h^{l-1} + b^{l}_i \). See~\citep{fischetti2018deep, anderson2019strong} for details.
After formulating the ReLU, if we relax the binary constraint~\eqref{eq:v_bounds_1} on \( z^{l}_i\) to \( [0, 1] \), we obtain a linear programming problem which is  easier and faster to solve. Furthermore, the quality (tightness) of such relaxation highly depends on the choice of tight upper and lower bounds, \( U^l_i, L^l_i\). In fact, the determination of tight bounds reduces the search space and hence, the solving time.

\section{MIP Constraints}\label{sec:mip_formulation}

In what follows, we adapt the MIP constraints~\eqref{eq:relu_constraints} to quantify neuron importance, and we describe the computation of the bounds \(L_i^l\) and \(U_i^l\). Our goal is to compute importance scores for all layers in the model in an integrated fashion, as~\citet{yu2018nisp} have shown to lead to better predictive accuracy than layer by layer.

\subsection{ReLU Layers}\label{sec:mip_relu_formulation}
In ReLU activated layers, we keep the previously introduced binary variables \(z^{l}_i\), and continuous variables \(h_i^l\). Additionally, we create the continuous decision variables \(s^l_i \in \left[0,1\right]\) representing neuron \(i\) importance score in layer \(l\). In this way, we modified the ReLU constraints~\eqref{eq:relu_constraints} by adding the neuron importance decision variable \(s^l_i\) to constraints~\eqref{eq:z_i_upper} and~\eqref{eq:z_i_lower}:
\begin{subequations}\label{eq:relu_constraints_neuron_score}
    \begin{alignat}{2}
        h^{l}_{i} + (1- z^{l}_i) L^{l}_i & \leq w_i^l  h^{l-1} + b^{l}_i - (1- s^{l}_i) \max{(U^{l}_i, 0)} \label{eq:z_i_upper_neuron_score}    \\
        h^{l}_{i}                        & \geq  w_i^l h^{l-1} + b^{l}_i - (1- s^{l}_i) \max{(U^{l}_i, 0)}.   \label{eq:z_i_lower_neuron_score}
    \end{alignat}
\end{subequations}
In~\eqref{eq:relu_constraints_neuron_score}, when neuron \(i\) is activated due to the input \(h^{l-1}\), i.e. \(z_i^l=1\), \(h^l_i\) is equal to the right-hand-side of those constraints. This value can be directly decreased by reducing the neuron importance \(s_i^l\). When neuron \(i\) is non-active, i.e.  \(z_i^l=0\), constraint~\eqref{eq:z_i_lower_neuron_score} becomes irrelevant as its right-hand-side is negative. This fact together with constraints~\eqref{eq:always_positive} and~\eqref{eq:upper_bound_relu}, imply that \(h^l_i\) is zero. Now, we claim that constraint~\eqref{eq:z_i_upper_neuron_score} allows \(s_i^l\) to be zero if that neuron is indeed non-important, i.e., for all possible input data points, neuron \(i\) is not activated. This claim can be shown through the following observations. Note that decisions \(h\) and \(z\) must be replicated for each input data point \(x\) as they present the propagation of \(x\) over the neural network. On the other hand, \(s\) evaluates the importance of each neuron for the main learning task and thus, it must be the same for all data input points. Thus, the key ingredients are the bounds \(L_i^l\) and \(U_i^l\) that are computed for each input data point, as explained in Sec.~\ref{sec:bound_prop}. In this way, if \(U_i^l\) is non-positive, \(s_i^l\) can be zero without interfering with the constraints~\eqref{eq:relu_constraints_neuron_score}. The latter is enforced by the objective function derived in Sec.~\ref{sec:mip_ojb}. We note that this MIP formulation can naturally be extended to convolutional layers converted to matrix multiplication using toeplitz matrix \citep{gray2000toeplitz} and with an importance score associated with each feature map. We refer the reader to the appendix for a detailed explanation.

\subsection{Bounds Propagation}\label{sec:bound_prop}
In the previous MIP formulation, we assumed  a large upper bound \(U^l_i\) and a small lower bound \(L^l_i\). However, using large bounds may lead to long computational times and a lost on the freedom to reduce the importance score as discussed above.
In order to overcome these issues, we tailor these bounds accordingly with their respective input point \(x\) by considering small perturbations on its value:
\begin{subequations}\label{eq:bounded_constraints}
    \begin{alignat}{2}
        L^0                 & = x - \epsilon                                              \\
        U^0                 & = x + \epsilon                                              \\
        L^l                 & = \myMatrix{W^{(l-)}} U^{l-1} + \myMatrix{W^{(l+)}} L^{l-1} \\
        U^l                 & = \myMatrix{W^{(l+)}} U^{l-1} + \myMatrix{W^{(l-)}} L^{l-1} \\
        \myMatrix{W^{(l-)}} & \triangleq \min{(\myMatrix{W^{(l)}}, 0)}                    \\
        \myMatrix{W^{(l+)}} & \triangleq \max{(\myMatrix{W^{(l)}}, 0)}.
    \end{alignat}
\end{subequations}
Propagating the initial bounds of the input  data points throughout the trained model will create the desired bound using simple arithmetic interval~\citep{moore2009introduction}.
The obtained bounds are tight, narrowing the space of feasible solutions.

\section{MIP Objectives}\label{sec:mip_ojb}
The aim for the proposed framework is to sparsify non-critical neurons without reducing the predictive accuracy of the pruned ANN\@
To this end, we combine two optimization objectives.

Our first objective is to maximize the set of neurons sparsified from the trained ANN\@. Let \(n\) be the number of layers, \(N^l\) the number of neurons at layer \(l\), and \(I^{l} =  \sum_{i = 1}^{N^l} (s^l_i -2)\) be the sum of neuron importance scores at layer \(l\) with \(s_i^l\)  scaled down to the range \([-2, -1]\). We refer the reader to the appendix~\ref{app:sparsification} for re-scaling experiments.

In order to create a relation between neurons' importance score in different layers, our objective becomes the maximization on the amount of neurons sparsified from the \(n-1\) layers with higher score \(I^l\).
Hence, we denote \(A = \{I^l : l=1,\ldots,n\}\) and formulate the sparsity loss as
\begin{equation}\label{eq:sparsification_objective}
    \text{sparsity} = \frac{\displaystyle \max_{A^{'} \subset A, |A^{'}| = (n-1)} \sum_{I \in A^{'}} I}{\sum_{l=1}^{n} \vert N^l \vert}.
\end{equation}
Here, the objective is to maximize the number of non-critical neurons at each layer compared to other layers in the trained neural model. Note that only the \(n-1\) layers with the largest importance score will weight in the objective, allowing to reduce the pruning effort on some layer that will naturally have low scores.
The sparsity quantification is then normalized by the total number of neurons.

Our second objective is to minimize the loss of \textit{important} information due to the sparsification of the trained neural model.
Additionally, we aim for this minimization to be done without relying on the values of the logits, which are closely correlated with neurons pruned at each layer. Otherwise, this would drive the MIP to simply give a full score of \(1\) to all neurons in order to keep the same output logit value. Instead, we formulate this optimization objective using the marginal softmax as proposed in \citep{gimpel2010softmax}. Using marginal softmax allows the solver to focus on minimizing the misclassification error without relying on logit values. Marginal softmax loss avoids putting a large weight on logits coming from the trained neural network and predicted logits from decision vector \(h^n\) computed by the MIP\@. On the other hand, in the proposed marginal softmax loss, the label having the highest logit value is the one optimized regardless its value. Formally, we write the objective
\begin{equation} \label{eq:softmax}
    \text{softmax} = \sum_{i = 1}^{N^n}  \log\left[\sum_c \exp(h^n_{i, c}) \right]  - \sum_{i =1}^{N^n} \sum_c Y_{i, c} h^n_{i, c},
\end{equation}
where index \(c\) stands for the class label. The used marginal softmax objective keeps the correct predictions of the trained model for the input batch of images \(x\) having one hot encoded labels \(Y\) without considering the logit value.

Finally, we combine the two objectives to formulate the multi-objective loss
\begin{equation} \label{eq:final_objective}
    \text{loss} = \text{sparsity} + \lambda \cdot \text{softmax}
\end{equation}
as a weighted sum of sparsification regularizer and marginal softmax, as proposed by \citet{ehrgott2005multicriteria}.
Our experiments revealed that \(\lambda=5\) generally provides the right trade-off between our two objectives; see the appendix for experiments with value of \(\lambda \).

\section{Empirical Results}\label{sec:experiments}
We first show in  Sec.~\ref{sec:mip_rob} the robustness of our proposed formulation to different input data points and different convergence levels of a neural network.  %
Next, in Sec.~\ref{sec:comparison_rand_critical}, we validate empirically that the computed neuron importance scores are meaningful, i.e.\ it is crucial to guide the pruning accordingly with the determined scores.
In Sec.~\ref{sec:generalization_dif_ds}, we proceed with experiments  to show that sub-networks generated by our approach on a specific initialization can be transferred to another dataset with marginal loss in accuracy (lottery hypothesis).
Finally, in Sec.~\ref{sec:comparisonSNIP}, we compare our masking methodology to \citep{yu2018nisp}, a framework used to compute connections sensitivity, and to create a sparsified sub-network based on the input dataset and model initialization. Before advancing to our results, we detail our experimental settings\footnote{The code can be found here: \url{https://github.com/chair-dsgt/mip-for-ann}.}.

\subsection{Experimental Setting} \label{sec:experimental_settings}
\paragraph{Architectures and Training}  We used a simple fully connected 3-layer ANN (FC-3) model, with 300+100 hidden units, from \citep{lecun1998gradient}, and another simple fully connected 4-layer ANN (FC-4) model, with 200+100+100 hidden units.  In addition, we used convolutional LeNet-5~\citep{lecun1998gradient} consisting of two sets of convolutional and average pooling layers, followed by a flattening convolutional layer, then two fully-connected layers. The largest architecture investigated was VGG-16~\citep{simonyan2014very} consisting of a stack of convolutional (conv.) layers with a very small receptive field: \( 3 \times 3 \). The VGG-16 was adapted for CIFAR-10~\citep{krizhevsky2009learning} having 2 fully connected layers of size 512 and average pooling instead of max pooling. Each of these models was trained 3 times with different initialization.

All models were trained for 30 epochs using RMSprop~\citep{tieleman2012lecture} optimizer with 1e-3 learning rate for MNIST and Fashion MNIST\@. Lenet 5~\citep{lecun1998gradient} on CIFAR-10 was trained using SGD optimizer with learning rate 1e-2 and 256 epochs. VGG-16~\citep{simonyan2014very}  on CIFAR-10 was trained using Adam~\citep{kingma2015adam} with 1e-2 learning rate for 30 epochs. Decoupled greedy learning~\citep{belilovsky2019decoupled} was used to train each VGG-16's layer using a small auxiliary network, and the neuron importance score was computed independently on each auxiliary network; then we fine-tuned the generated masks for 1 epoch to propagate error across them. Decoupled training of each layer allowed us to represent deep models using the MIP formulation and to parallelize the computation per layer; see appendix for details about decoupled greedy learning. The hyper parameters were tuned on the validation set's accuracy.
All images were resized to 32 by 32 and converted to 3 channels to generalize the pruned network across different datasets.

\paragraph{MIP and Pruning Policy} Using all the training set as input to the MIP solver is intractable. Hence, we only use a subset of the data points to approximate the neuron importance score. Representing classes with a subset of the data points would give us an under estimation of the score, i.e., neurons will look less critical than they really are. To that extent, the selected subset of data points must be carefully chosen. Whenever we computed neuron scores for a trained model, we fed the MIP with a balanced set of images, each representing a class of the classification task. The aim was to avoid that the determined importance scores lead to pruning neurons (features) critical to a class represented by  fewer images as input to the MIP.  We used \( \lambda=5 \) in the MIP objective function~\eqref{eq:final_objective}; see appendix for experiments with the value of \( \lambda \).
The proposed framework, recall Figure~\ref{fig:method_diagram}, computes the importance score of each neuron, and with a small tuned threshold  based on the network's architecture, we  masked (pruned) non-critical neurons with a score lower than it.

\paragraph{Computational Environment} The experiments were performed in an Intel(R) Xeon(R) CPU @ 2.30GHz with 12 GB RAM and Tesla k80 using Mosek 9.1.11 \citep{mosek2010mosek} solver on top of CVXPY \citep{cvxpy_rewriting, cvxpy} and PyTorch 1.3.1 \citep{pytorch_2019}.

\subsection{MIP Robustness}\label{sec:mip_rob}
\begin{figure}[ht]
    \centering
    \begin{subfigure}[t]{0.47\textwidth}
        \centering
        \fbox{\includegraphics[width=0.47\textwidth ,height=2.5cm]{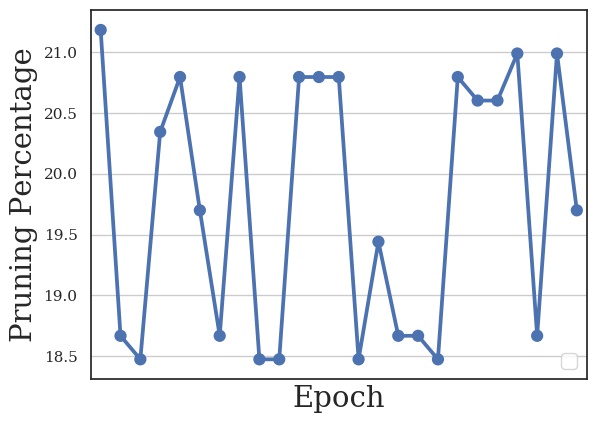}\includegraphics[width=0.5\columnwidth,height=2.5cm]{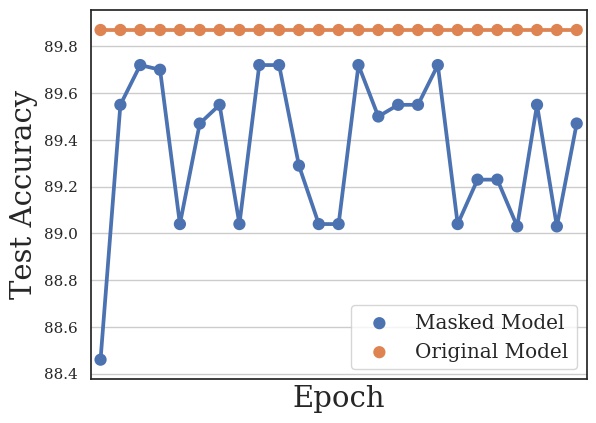}}
        \caption{Effect of changing validation set of input images.}
        \label{fig:verify_fmnist}
    \end{subfigure}
    \begin{subfigure}[t]{0.47\textwidth}
        \centering
        \fbox{\includegraphics[width=0.47\columnwidth,height=2.5cm]{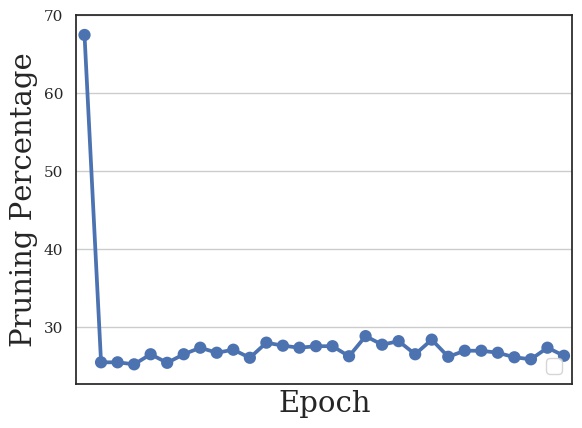}\includegraphics[width=0.5\columnwidth,height=2.5cm]{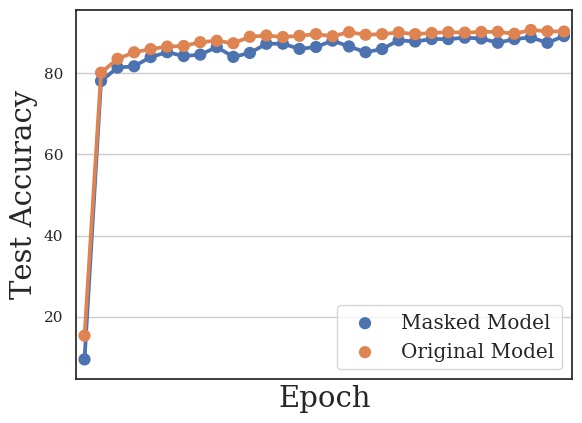}}
        \caption{Evolution of the computed masked subnetwork during model training.}
        \label{fig:pruning_incremental_prune_fmnist}
    \end{subfigure}
\end{figure}

\begin{wraptable}[15]{R}{0.5\linewidth}
    \caption{Comparing test accuracy of LeNet-5 on imbalanced independent class by class (IMIDP.), balanced  independent class by class (IDP.) and simultaneously all classes (SIM) with 0.01 threshold, and \(\lambda=1 \).}
    \label{tab:compare_parallel_class_imb}
    \vspace{-1pt}
    \centering
    \begin{small}
        \begin{sc}
            \resizebox{0.45\textwidth}{!}{%
                \begin{tabular}{lcr}
                    {}         & MNIST                & Fashion-MNIST        \\
                    \hline
                    Ref.       & $98.8\% \pm 0.09$    & $89.5\% \pm 0.3$     \\
                    \hline
                    IDP.       & $98.6\% \pm  0.15 $  & $87.3\% \pm  0.3 $   \\
                    Prune (\%) & $19.8\%  \pm 0.18 $  & $21.8\%  \pm 0.5 $   \\
                    \hline
                    IMIDP.     & $98.6\%  \pm 0.1 $   & $88\%  \pm 0.1 $     \\
                    Prune (\%) & $15\% \pm  0.1 $     & $ 18.1\%  \pm  0.3 $ \\
                    \hline
                    SIM.       & $98.4\%  \pm 0.3 $   & $87.9\%  \pm 0.1$    \\
                    Prune (\%) & $13.2\%  \pm  0.42 $ & $18.8\%  \pm  1.3 $  \\
                    \hline
                \end{tabular}
            }
        \end{sc}
    \end{small}
\end{wraptable}
We examine the robustness of our formulation against different batches of input images fed into the MIP\@. Namely, we used 25 randomly sampled balanced images from the validation set. Figure~\ref{fig:verify_fmnist} shows that changing the input images used  by the MIP to compute neuron importance scores  resulted in marginal changes in the test accuracy between different batches. We remark that the input batches may contain images that were misclassified by the neural network. In this case, the MIP tries to use the score \(s\) to achieve the true label, explaining the variations on the pruning percentage. Indeed, as discussed in appendix for the choice of \(\lambda \), the marginal fluctuations of these results depend on the accuracy of the input batch used in the MIP\@. Additionally, we demonstrate empirically that we can parallelize the computation of neuron scores per class as it shows comparable results to feeding all data points to the MIP at once; see Table~\ref{tab:compare_parallel_class_imb} and appendix for extensive experiments. For those experiments, we sampled a random number of images per class, and then we took the average of the computed neuron importance scores from solving the MIP on each class. The obtained sub-networks were compared to solving the MIP with 1 image per class.  We achieved comparable results in terms of test accuracy and pruning percentage. In brief, our method is empirically shown to be scalable and that class contribution can be decoupled without deteriorating the approximation of neuron scores and thus, the performance of our methodology.

To conclude on the robustness of the scores computed based on the input points used in the MIP, we show in Table~\ref{tab:compare_parallel_class_imb} that our formulation is robust even when an imbalanced number of data points per class is used in the MIP\@.

Finally, we also tested the robustness of our approach along the evolution of neuron importance scores during training between epochs. To this end, we computed neuron importance scores after each epoch jointly with the respective pruning. As shown in Figure~\ref{fig:pruning_incremental_prune_fmnist},  our proposed formulation can identify non-critical neurons in the network before the model's convergence.

\subsection{Comparison to Random and Critical Pruning}\label{sec:comparison_rand_critical}
We started by training a reference model (REF.) using the training parameters in Sec.~\ref{sec:experimental_settings}.
After training and evaluating the reference model on the test set, we fed an input batch of images from the validation set to the MIP.
Then, the MIP solver computed the neuron importance scores based on those input images. In our experimental setup, by taking advantage of the conclusions from the previous section, we used \(10\)  images, each representing a class.

\begin{table}[ht]
    \caption{Pruning results on fully connected (FC-3, FC-4) and convolutional (Lenet-5, VGG-16) network architectures using three different datasets. We compare the test accuracy between the unpruned reference network (REF.), randomly pruned model (RP.), model pruned based on critical neurons selected by the MIP (CP.) and our non-critical pruning approach with (OURS + FT) and without (OURS) fine-tuning for 1 epoch.}
    \label{tab:compare_random_critical_noncritical_methods}
    \vspace{3pt}
    \centering
    \begin{small}
        \begin{sc}
            \resizebox{\columnwidth}{!}{%
            \begin{tabular}{@{}ccccccccc@{}}
                \toprule
                {}                             & {}                         & \multicolumn{1}{l}{Ref.}                 & \multicolumn{1}{l}{RP.}                & \multicolumn{1}{l}{CP.}                  & \multicolumn{1}{l}{Ours} & \multicolumn{1}{l}{Ours + FT}          & \multicolumn{1}{l}{Prune (\%)}         & \multicolumn{1}{l}{threshold} \\ \midrule
                \multirow{3}{*}{MNIST}         & FC-3                       & \( 98.1\% \pm 0.1 \)                     & \( 83.6\% \pm 4.6 \)                   & \( 44.5\% \pm 7.2 \)                     & \boldsymbol{$95.9\% \pm 0.87$}      & \boldsymbol{$97.8 \pm 0.2$}                       & \( 44.5\% \pm 7.2 \)                   & \( 0.1 \)                     \\
                                               & FC-4                       & \( 97.9\% \pm 0.1 \)                     & \( 77.1 \% \pm 4.8 \)                  & \( 50\% \pm 15.8 \)                      & \boldsymbol{$96.6\% \pm 0.4$}       & \boldsymbol{$97.6\% \pm 0.01$}                    & \( 42.9\% \pm 4.5 \)                   & \( 0.1 \)                     \\
                                               & LeNet-5                    & \(98.9\% \pm 0.1\)                       & \( 56.9\% \pm 36.2 \)                  & \( 38.6\% \pm 40.8 \)                    & \boldsymbol{$98.7\% \pm 0.1$}       & \boldsymbol{$98.9\% \pm 0.04$}                    & \( 17.2\% \pm 2.4 \)                   & \( 0.2 \)                     \\
                \midrule
                \multirow{3}{*}{Fashion-MNIST} & FC-3                       & \( 87.7\% \pm 0.6 \)                     & \( 35.3\% \pm 6.9  \)                  & \( 11.7\% \pm 1.2 \)                     & \boldsymbol{$80\% \pm 2.7$}         & \boldsymbol{$88.1\% \pm 0.2$}                     & \( 68\% \pm 1.4 \)                     & \( 0.1 \)                     \\
                                               & FC-4                       & \( 88.9\% \pm 0.1 \)                     & \( 38.3\% \pm 4.7 \)                   & \( 16.6\% \pm 4.1 \)                     & \boldsymbol{$86.9\% \pm 0.7$}       & \boldsymbol{$88\% \pm 0.03$}                      & \( 60.8\% \pm 3.2 \)                   & \( 0.1 \)                     \\
                                               & LeNet-5                    & \( 89.7\% \pm 0.2\)                      & \( 33\%  \pm 24.3 \)                   & \( 28.6\% \pm 26.3 \)                    & \boldsymbol{$87.7\% \pm 2.2$}       & \boldsymbol{$89.8\% \pm 0.4$}                     & \( 17.8\% \pm 2.1 \)                   & \( 0.2 \)                     \\
                \midrule
                \multirow{2}{*}{CIFAR-10}      & LeNet-5                    & \( 72.2\% \pm 0.2 \)                     & \( 50.1 \% \pm 5.6 \)                  & \( 27.5\% \pm 1.7 \)                     & \boldsymbol{$67.7\% \pm 2.2$}       & \boldsymbol{$68.6\% \pm 1.4$}                     & \( 9.9\% \pm 1.4 \)                    & \(0.3\)                       \\
                                               & \multicolumn{1}{r}{VGG-16} & \multicolumn{1}{r}{\( 83.9\% \pm 0.4 \)} & \multicolumn{1}{r}{\( 85\% \pm 0.4 \)} & \multicolumn{1}{r}{\( 83.3\% \pm 0.3 \)} & \multicolumn{1}{r}{N/A}             & \multicolumn{1}{r}{\boldsymbol{$85.3\% \pm 0.2$}} & \multicolumn{1}{r}{\( 36\% \pm 1.1 \)} & \( 0.3 \)                     \\ \bottomrule
            \end{tabular}
            }
        \end{sc}
    \end{small}
\end{table}
In order to validate our pruning policy guided by the computed importance scores,  we created different sub-networks of the reference model, where the same number of neurons is removed in each layer, thus allowing a fair comparison among them. These sub-networks were obtained through  different procedures: non-critical (our methodology), critical and randomly pruned neurons. For VGG-16 experiments, an extra fine-tuning step for 1 epoch is performed on all generated sub-networks. Although we pruned the same number of neurons, which accordingly with~\citep{liu2018rethinking} should result in similar performances, Table~\ref{tab:compare_random_critical_noncritical_methods} shows that pruning non-critical neurons results in marginal loss and gives better performance. On the other hand, we observe a significant drop on the test accuracy when critical or a random set of neurons are removed compared with the reference model. If we fine-tune for just 1 epoch the sub-network obtained through our method, the model's accuracy can surpass the reference model. This is due to the fact that the MIP, while computing neuron scores, is solving its marginal softmax~\eqref{eq:softmax} on true labels.

\subsection{Generalization Between Different Datasets}\label{sec:generalization_dif_ds}

\begin{wraptable}[10]{R}{0.7\linewidth}
  \caption{Cross-dataset generalization: sub-network masking is computed on source dataset (\(d_1\)) and then applied to target dataset (\(d_2\)) by retraining with the same early initialization. Test accuracies are presented for masked and unmasked (REF.) networks on \(d_2\), as well as pruning percentage.}
  \label{tab:large_threshold_generalize_fashion_mnist_1}
  \vskip -0.08in
  \vspace{-2pt}
  \begin{center}
    \begin{small}
      \begin{sc}
        \resizebox{0.7\textwidth}{!}{%
          \begin{tabular}{lcccccr}
            \toprule
            Model                    & Source dataset \(d_1\)    & Target dataset \(d_2\) & REF. Acc.          & Masked Acc.        & Pruning (\%)                         & {} \\
            \midrule
            \multirow{2}{*}{LeNet-5} & \multirow{2}{*}{Mnist}    & Fashion MNIST          & \(89.7\% \pm 0.3\) & \(89.2\% \pm 0.5\) & \multirow{2}{*}{ \(16.2\% \pm 0.2\)}      \\
            {}                       & {}                        & CIFAR-10               & \(72.2\% \pm 0.2\) & \(68.1\% \pm 2.5\) & {}                                        \\
            \midrule
            \multirow{2}{*}{VGG-16}  & \multirow{2}{*}{CIFAR-10} & MNIST                  & \(99.1\%\pm0.1\)   & \(99.4\%\pm0.1\)   & \multirow{2}{*}{\(36\% \pm 1.1\)}         \\
            {}                       & {}                        & Fashion-Mnist          & \(92.3\% \pm 0.4\) & \(92.1\% \pm 0.6\) & {}                                        \\
            \bottomrule
          \end{tabular}}
      \end{sc}
    \end{small}
  \end{center}
  \vskip -0.1in
\end{wraptable}

In this experiment, we train the model on a dataset \( d_1 \), and we create a masked neural model using our approach.
After creating the masked model, we restart it to its original initialization. Finally, the new masked model is re-trained on another dataset \( d_2 \), and its generalization is analyzed.

Table~\ref{tab:large_threshold_generalize_fashion_mnist_1} displays our experiments and respective results. When we compare generalization results to pruning using our approach on Fashion-MNIST and CIFAR-10, we discover that computing the critical sub-network LeNet-5 architecture on MNIST, is creating a more sparse sub-network with test accuracy better than zero-shot pruning without fine-tuning using our approach, and comparable accuracy with the original ANN\@.
This behavior is happening because the solver is optimizing on a batch of images that are classified correctly with high confidence from the trained model.
Furthermore, computing the critical VGG-16 sub-network architecture on CIFAR-10 using decoupled greedy learning~\citep{belilovsky2019decoupled} generalizes well to Fashion-MNIST and MNIST\@.

\subsection{Comparison to SNIP}\label{sec:comparisonSNIP}

Our proposed framework can be viewed as a compression technique of over-parameterized neural models. In what follows, we compare it to the state-of-the-art framework: SNIP~\citep{lee2018snip}.
SNIP creates the sparse model before training the neural model by computing the sensitivity of connections. This allows the identification of the important connections.
In our methodology, we exclusively identify the importance of neurons and prune all the connections of non-important ones. On the other hand, SNIP only focus on pruning neurons' connections. Moreover, we highlight that SNIP can only compute connection's sensitivity on ANN's initialization. As for a trained ANN, the magnitude of the derivatives with respect to the loss function was optimized during the training, making SNIP more keen to keep all the parameters. On the other hand, our framework can work on different convergence levels as shown in Sec.~\ref{sec:mip_rob}. Furthermore, the connection's sensitivity computed is only network and dataset specific, thus the computed connection sensitivity for a single connection does not give a meaningful signal about its importance to the task at hand, but needs to be compared to the sensitivity of other connections.

In order to bridge the differences between the two methods, and provide a fair comparison in equivalent settings, we make slight adjustments.  We compute neuron importance scores on the model's initialization\footnote{Remark: we used \( \lambda=1 \) and pruning threshold \( 0.2 \) and kept ratio \( 0.45 \) for SNIP\@. Training procedures as in Section~\ref{sec:experimental_settings}.}. We used only 10 images as an input to the MIP corresponding to 10 different classes, and 128 images as input to SNIP, as in the associated paper~\citep{lee2018snip}. Our algorithm was able to prune neurons from fully connected and convolutional layers of LeNet-5. After creating the sparse network using both SNIP and our methodology, we  trained them on Fashion-MNIST dataset. The difference between SNIP (\( 88.8\% \pm 0.6 \))  and our approach (\( 88.7\% \pm 0.5 \)) was marginal in terms of test accuracy. SNIP pruned \( 55\% \) of the ANN's parameters and our approach \( 58.4\% \).

In brief, we remark that the adjustments made to SNIP and our framework in the previous experiments are for the purpose of comparison, while the main purpose of our method is to allow optimization at any stage (before, during, or after training). In the specific case of optimizing over initialization and discarding entire neurons based on connection sensitivity, the SNIP approach may have some advantages, notably in scalability for deep architectures. However, it also has some limitations, as discussed before.

\section{Conclusion}\label{sec:discussion}
We proposed a mixed integer program to compute neuron importance scores in ReLU-based deep neural networks. Our contributions focus here on providing scalable computation of importance scores in fully connected and convolutional layers. We presented results showing these scores can be effectively used to prune unimportant parts of the network without significantly affecting its main task (e.g., showing small or negligible drop in classification accuracy). Further, our results indicate this approach allows automatic construction of efficient sub-networks that can be transferred and retrained on different datasets. The presented model introduces one of the first steps in understanding which components in a neural network are critical for its model capacity to perform a given task, which can have further impact in future work beyond the pruning applications presented here.

\begin{ack}
\noindent This work was partially funded by: IVADO (l'institut de valorisation des donn\'{e}es) [\emph{G.W.}, \emph{M.C.}]; NIH grant R01GM135929 [\emph{G.W.}]; FRQ-IVADO Research Chair in Data Science for Combinatorial Game Theory, and NSERC grant 2019-04557 [\emph{M.C.}].
\end{ack}

\bibliographystyle{unsrtnat}
\bibliography{neurips_paper.bib}

\clearpage
\appendix

\section{Appendix}
\section{MIP formulations}\label{app:mip}

In Appendix~\ref{app:convolutional}, details on the MIP constraints for convolutional layers are provided. Appendix~\ref{app:pooling}  explains the formulation used to represent pooling layers. Appendix~\ref{app:lambada} discusses the parameter \(\lambda\) in the objective function~\eqref{eq:final_objective} guiding the computation of neuron importance scores.

\subsection{MIP for convolutional layers}\label{app:convolutional}

We convert the convolutional feature map to a toeplitz matrix and the input image to a vector. This allow us to use simple matrix multiplication which is computationally efficient. Moreover, we can represent the convolutional layer using the same formulation of fully connected layers presented in Sec.~\ref{sec:mip_formulation}.

\textbf{Toeplitz Matrix} is a matrix in which each value is along the main diagonal and sub diagonals are constant.
So given a sequence \(a_n\), we can create a Toeplitz matrix by  putting the sequence in the first column of the matrix and then shifting it by one entry  in the following columns:

\begin{equation}
    \begin{pmatrix}
        a_0       & a_{-1} & a_{-2} & \cdots & \cdots & \cdots & \cdots & a_{-(N-1)} \\
        a_1       & a_0    & a_{-1} & a_{-2} &        &        &        & \vdots     \\
        a_2       & a_1    & a_0    & a_{-1} & \ddots &        &        & \vdots     \\
        \vdots    & a_2    & \ddots & \ddots & \ddots & \ddots &        & \vdots     \\
        \vdots    &        & \ddots & \ddots & \ddots & \ddots & a_{-2} & \vdots     \\
        \vdots    &        &        & \ddots & a_1    & a_0    & a_{-1} & a_{-2}     \\
        \vdots    &        &        &        & a_2    & a_1    & a_0    & a_{-1}     \\
        a_{(N-1)} & \cdots & \cdots & \cdots & \cdots & a_2    & a_1    & a_0        \\
    \end{pmatrix}.
\end{equation}
\vspace{10mm}

\textbf{Feature maps} are flipped   and   then converted to a matrix. The computed matrix when multiplied by the vectorized input image will compute the fully convolutional output. For padded convolution we use only parts of the output of the full convolution, for strided convolutions we used sum of 1 strided convolution as proposed by~\citet{brosch2015efficient}. First, we pad zeros to the top and right of the input feature map to become same size as the output of the full convolution. Next, we create a toeplitz matrix for each row of the zero padded feature map. Finally, we arrange these small toeplitz matrices in a big doubly blocked toeplitz matrix. Each small toeplitz matrix is arranged in the doubly toeplitz matrix in the same way a toeplitz matrix is created from input sequence with each small matrix as an element of the sequence.

\subsection{Pooling Layers}\label{app:pooling}

We represent both average and max pooling on multi-input units in our MIP formulation. Pooling layers are used to reduce spatial representation of input image by applying an arithmetic operation on each feature map of the previous layer.

\textbf{Avg Pooling} layer  applies the average operation on each feature map of the previous layer. This operation is linear and thus, it can directly be included in the MIP constraints:
\begin{equation}\label{eq:avg_pooling}
    h^{l+1} =  \text{AvgPool}(h^l_1,\cdots,h^{l}_{N^l} ) = \frac{1}{N^l} \sum_{i=1}^{N^l} h^l_i.
\end{equation}

\textbf{Max Pooling} takes the maximum of each feature map of the previous layer:
\begin{equation}\label{eq:max_pooling}
    h^{l+1} =  \text{MaxPool}(h^l_1,\cdots,h^l_{N^l} ) =  \text{max}\{h^l_1, \cdots, h^l_{N^l}\}  .
\end{equation}
This operation can be expressed by introducing a set of binary variables $m_1, \cdots ,m_{N^l}$, where $m_i=1$ implies  $x=\text{MaxPool}(h^l_1,\cdots,h^l_{N^l} )$:

\begin{subequations}\label{eq:max_pooling_constr}
    \begin{align}
        \sum_{i=1}^{N^l} m_i  = 1 \ \ \ \ \ \ \ \ \ \ \ \ \ \ \ \ \ \ \ \ \ \ \ \ \\
        \begin{rcases}
            x \ge h^l_i,  \\
            x \le h^l_im_i+U_i(1-m_i)    \\
            m_i \in \{0,1\}
        \end{rcases}    i=1,\cdots,N^l.
    \end{align}
\end{subequations}

\subsection{Choice of Lambda}\label{app:lambada}
\begin{wrapfigure}[12]{R}{0.64\linewidth}
    \vspace{-10pt}
    \centering
    \fbox{\includegraphics[width=0.3\textwidth]{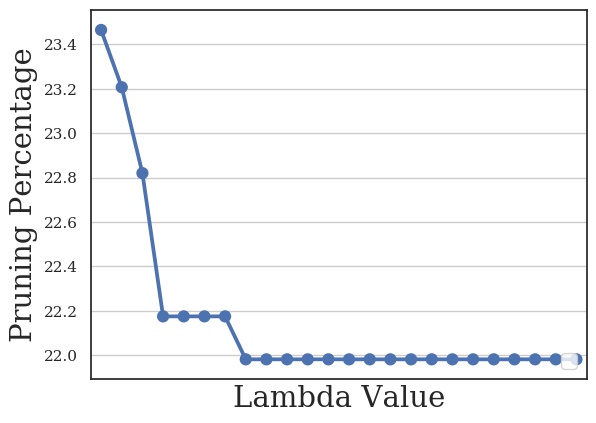} \includegraphics[width=0.3\textwidth]{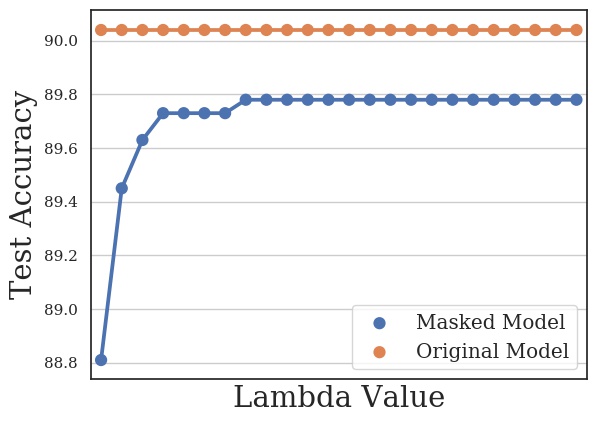}}
    \caption{Effect of changing value of $\lambda$ when pruning LeNet model trained on Fashion MNIST.}
    \label{fig:verify_lambda}
\end{wrapfigure}

Note that our objective function~\eqref{eq:final_objective} is implicitly using a Lagrangian relaxation, where \(\lambda\geq 0\) is the Lagrange multiplier. In fact, one would like to control the loss on accuracy~\eqref{eq:softmax} by imposing the constraint \(\text{softmax}(h) \leq \epsilon\) for a very small \()\epsilon\), or even to avoid any loss via \(\epsilon=0\). However, this would introduce a nonlinear constraint which would be hard to handle. Thus, for tractability purposes we follow a Lagrangian relaxation on this constraint, and penalize the objective whenever \(\text{softmax}(h)\) is positive. Accordingly with the weak (Lagrangian) duality theorem, the objective~\eqref{eq:final_objective} is always a lower bound to the problem where we minimize sparsity and a bound on the accuracy loss is imposed. Furthermore, the problem of finding the best \(\lambda\) for the Lagrangian relaxation, formulated as
\begin{equation}\label{eq:BestLagrangian}
    \max_{\lambda\geq 0} \min\{ \text{sparsity} + \lambda \cdot \text{softmax}\},
\end{equation}
has the well-known property of being concave, which in our experiments revealed to be easily determined\footnote{We remark that if the trained model has misclassifications, there is no guarantee that problem~\eqref{eq:BestLagrangian} is concave.}. We note that the value of \(\lambda \) introduces a trade off between pruning more neurons and the predictive capacity of the model. For example, increasing the value of \(\lambda \) would result on pruning fewer neurons, as shown in Figure~\ref{fig:verify_lambda}, while the accuracy on the test set would increase.

\subsection{Re-scaling of MIP Sparsification Objective}\label{app:sparsification}

\begin{table}[!ht]
    \caption{Importance of re-scaling sparsification objective to prune more neurons shown empirically on LeNet-5 model using threshold 0.05, by comparing accuracy on test set between reference model (Ref.), and pruned model (Masked).}
    \label{tab:compare_objectives}
    \begin{center}
        \begin{small}
            \begin{sc}
                \resizebox{\columnwidth}{!}{%
                    \begin{tabular}{lllll}
                        Dataset                        & Objective      & Ref. Acc.                              & Masked Acc.                      & Pruning Percentage (\%)          \\
                        \hline
                        \multirow{3}{*}{MNIST}         & \( s^l_i -2 \) & \multirow{3}{*}{\( 98.9 \% \pm 0.1 \)} & \boldsymbol{ $98.7 \% \pm 0.1 $} & \boldsymbol{ $ 13.2\% \pm 2.9 $} \\
                        {}                             & \( s^l_i -1 \) & {}                                     & \( 98.8 \% \pm 0.1 \)            & \( 9.6\% \pm 1.1 \)              \\
                        {}                             & \( s^l_i  \)   & {}                                     & \( 98.9 \% \pm 0.2 \)            & \( 8\% \pm 1.6 \)                \\
                        \midrule
                        \multirow{3}{*}{Fashion-MNIST} & \( s^l_i -2 \) & \multirow{3}{*}{\( 89.9 \% \pm 0.2 \)} & \boldsymbol{$89.1 \% \pm 0.3 $}  & \boldsymbol{$17.1\% \pm 1.2 $}   \\
                        {}                             & \( s^l_i -1 \) & {}                                     & \( 89.2 \% \pm 0.1 \)            & \( 17\% \pm 3.4 \)               \\
                        {}                             & \( s^l_i  \)   & {}                                     & \( 89 \% \pm 0.4 \)              & \( 10.8\% \pm 2.1 \)             \\
                        \bottomrule
                    \end{tabular}
                }
            \end{sc}
        \end{small}
    \end{center}
\end{table}
In Table~\ref{tab:compare_objectives}, we compare re-scaling the neuron importance score in the objective function to \( [-2, -1] \),  to \( [-1, 0] \) and no re-scaling \( [0, 1] \). This comparison shows empirically the importance of re-scaling the neuron importance score to optimize sparsification through neuron pruning. 

\section{Generalization comparison between SNIP and our approach}

\begin{table}[!ht]
    \caption{Cross-dataset generalization comparison between SNIP, with neurons having the lowest sum of connections' sensitivity pruned, and our framework (Ours), both applied on initialization, see Section~\ref{sec:generalization_dif_ds} for the generalization experiment description.}
    \label{tab:generalization_snip_oamip_compare}
    \begin{center}
        \begin{small}
            \begin{sc}
                \resizebox{0.7\textwidth}{!}{%
                    \begin{tabular}{lcccccr}
                        \toprule
                        Source dataset \( d_1 \) & Target dataset \( d_2 \)       & REF. Acc.                             & Method          & Masked Acc.                    & Pruning (\%)                    & {} \\
                        \midrule
                        \multirow{4}{*}{Mnist}   & \multirow{2}{*}{Fashion-MNIST} & \multirow{2}{*}{\( 89.7\% \pm 0.3 \)} & SNIP            & \( 85.8\% \pm 1.1 \)           & \( 53.5\% \pm 1.8 \)                 \\
                        {}                       & {}                             & {}                                    & Ours & \boldsymbol{ $ 88.5\%\pm0.3 $} & \boldsymbol{$ 59.1\% \pm 0.8 $}      \\
                        \cmidrule(lr){2-6}
                        {}                       & \multirow{2}{*}{CIFAR-10}      & \multirow{2}{*}{\( 72.2\% \pm 0.2 \)} & SNIP            & \( 53.5\% \pm 3.3 \)           & \( 53.5\% \pm 1.8 \)                 \\
                        {}                       & {}                             & {}                                    & Ours & \boldsymbol{ $63.6\% \pm 1.4$} & \boldsymbol{ $59.1\% \pm 0.8$}       \\
                        \bottomrule
                    \end{tabular}
                }
            \end{sc}
        \end{small}
    \end{center}
\end{table}

In Table~\ref{tab:generalization_snip_oamip_compare}, we show that our framework outperforms SNIP in terms of generalization. We adjusted SNIP to prune entire neurons based on the value of the sum of its connections' sensitivity, and our framework was also applied on ANN's initialization. When our framework is applied on the initialization, more neurons are pruned as the marginal softmax part of the objective discussed in Section~\ref{sec:mip_ojb} is weighting less (\(\lambda =1\)), driving the optimization to focus on model sparsification.

\section{Scalability improvements}\label{app:scalability}

Appendix~\ref{app:class_class} provides computational evidence on the parallelization of the importance scores computation. In Appendix~\ref{app:Decoupling}, we describe a methodology that aims at speed-up the computation of neuron importance scores by relying on decoupled greedy learning.

\subsection{MIP Class by Class}\label{app:class_class}
In this experiment, we show that the neuron importance scores can be approximated  by \emph{1)} solving for each class the MIP with only one data point from it, and \emph{2)}  taking the average of the computed scores for each neuron. Such procedure would speed-up our methodology for problems with a large number of classes. We compare the subnetworks obtained through this independent class by class approach (IDP.) and by feeding at once the same data points from all the classes to the MIP (SIM.) on Mnist and Fashion-Mnist using LeNet-5.

\begin{table}[ht]
    \caption{Comparing balanced independent class by class (IDP.) and simultaneously all classes (SIM.) with different thresholds using LeNet-5.}
    \label{tab:compare_parallel_class_sim_app}
    \vspace{3pt}
    \centering
    \begin{small}
        \begin{sc}
            \begin{tabular}{lccr}
                {}         & MNIST                & Fashion-MNIST       & threshold               \\
                \hline
                Ref.       & $98.8\% \pm 0.09$    & $89.5\% \pm 0.3$    & {}                      \\
                \hline
                IDP.       & $96.6 \pm  2.4 \%$   & $86.81 \pm  1.2 \%$ & \multirow{2}{*}{$0.1$}  \\
                Prune (\%) & $28.4  \pm 1.5 \%$   & $29.6  \pm 1.8 \%$  & {}                      \\
                \hline
                SIM.       & $98.5  \pm 0.28 \%$  & $88.7  \pm 0.4 \%$  & \multirow{2}{*}{$0.1$}  \\
                Prune (\%) & $16.5  \pm  0.5 \%$  & $18.9  \pm  1.4 \%$ & {}                      \\
                \hline
                \hline
                IDP.       & $98.6\% \pm  0.15 $  & $87.3\% \pm  0.3 $  & \multirow{2}{*}{$0.01$} \\
                Prune (\%) & $19.8\%  \pm 0.18 $  & $21.8\%  \pm 0.5 $  & {}                      \\
                \hline
                SIM.       & $98.4\%  \pm 0.3 $   & $87.9\%  \pm 0.1$   & \multirow{2}{*}{$0.01$} \\
                Prune (\%) & $13.2\%  \pm  0.42 $ & $18.8\%  \pm  1.3 $ & {}                      \\
                \hline
            \end{tabular}
        \end{sc}
    \end{small}
\end{table}

Table~\ref{tab:compare_parallel_class_sim_app} expands the results presented in Sec.~\ref{sec:mip_rob}, where we had discussed the comparable results between IDP. and SIM. when we use a small threshold 0.01. However, we can notice a difference between both of them when we use a threshold of 0.1. This difference comes from the fact that computing neuron importance scores on each class independently zeros out more neuron scores resulting in an average that leads more neurons to be pruned. %

\subsection{Decoupled Greedy Learning}\label{app:Decoupling}

We use decoupled greedy learning \citep{belilovsky2019decoupled} to parallelize learning of each layer by computing its gradients and using an auxiliary network attached to it. By using this procedure,  we have auxiliary networks of the deep neural network that represent subsets of layers thus allowing us to solve the MIP in sub-representations of the neural network.

\paragraph{Training procedure} We start by constructing auxiliary networks for each convolutional layer except the last convolutional layer that will be attached to the classifier part of the model. During the training each auxiliary network is optimized with a separate optimizer and the auxiliary network's output is used to predict the back-propagated gradients. Each sub-network's input is the output of the previous sub-network and the gradients will flow through only the current sub-network. In order to parallelize this operation a replay buffer of previous representations should be used to avoid waiting for output from previous sub-networks during training.

\paragraph{Auxiliary Network Architecture} We use a spatial averaging operation to construct a scalable auxiliary network applied to the output of the trained layer and to reduce the spatial resolution by a factor of 4, then applying one \(1 \times 1\) convolution with batchnorm~\citep{ioffe2015batch} followed by a reduction to \(2 \times 2\) and a one-layer MLP\@. The architecture used for the auxiliary network is smaller than the one mentioned in the paper leading to speed up the MIP solving time per layer.

\paragraph{MIP representation} After training each sub-network, we create a separate MIP formulation for each auxiliary network using its trained parameters and taking as input the output of the previous sub-network. This operation can be easily parallelized and each sub-network can be solved independently. Then, we take the computed neuron importance scores per layer and apply them to the main deep neural network. Since these layers were solved independently, we fine tune the network for one epoch to back-propagate the error across the network. The created sub-network can be generalized across different datasets and yields marginal loss.

\end{document}